%% file: iclr2026_conference.tex
\documentclass{article}
\usepackage{graphicx}
\usepackage{iclr2026_conference,times}

\input{math_commands.tex}

\usepackage{hyperref}
\usepackage{url}
\iclrfinalcopy

\title{FreqTrack: Frequency Learning based Vision Transformer for RGB-Event Object Tracking}

\author{Jinlin You\\
Dalian University of Technology\\
{\tt\small 2497066671@mail.dlut.edu.cn}
\and
Muyu Li\\
Dalian University of Technology\\
{\tt\small muyuli@dlut.edu.cn}
\and
Xudong Zhao\\
Dalian University of Technology\\
{\tt\small xudongzhao@dlut.edu.cn}
}

\begin{document}

\maketitle

\begin{abstract}
Existing single-modal RGB trackers often face performance bottlenecks in complex dynamic scenes, while the introduction of event sensors offers new potential for enhancing tracking capabilities. However, most current RGB-event fusion methods, primarily designed in the spatial domain using convolutional, Transformer, or Mamba architectures, fail to fully exploit the unique temporal response and high-frequency characteristics of event data. To address this, we1 propose FreqTrack, a frequency-aware RGBE tracking framework that establishes complementary inter-modal correlations through frequency-domain transformations for more robust feature fusion. We design a Spectral Enhancement Transformer (SET) layer that incorporates multi-head dynamic Fourier filtering to adaptively enhance and select frequency-domain features. Additionally, we develop a Wavelet Edge Refinement (WER) module, which leverages learnable wavelet transforms to explicitly extract multi-scale edge structures from event data, effectively improving modeling capability in high-speed and low-light scenarios. Extensive experiments on the COESOT and FE108 datasets demonstrate that FreqTrack achieves highly competitive performance, particularly attaining leading precision of 76.6\% on the COESOT benchmark, validating the effectiveness of frequency-domain modeling for RGBE tracking.
\end{abstract}

\section{Introduction}
\label{sec:intro}

Visual object tracking, as a core task in visual surveillance and autonomous systems, plays a critical role in scenarios requiring real-time localization and motion understanding. Although conventional RGB trackers~\citep{bhat2019learning,ye2022joint,yan2021learning,zheng2024odtrack} have achieved remarkable progress, they remain constrained by inherent limitations of frame-based imaging: motion blur under high-speed movement, limited dynamic range in challenging illumination, and intrinsic redundancy in sequential image processing. These issues become particularly pronounced in low-light or high-speed environments, where traditional RGB cameras struggle to capture precise temporal dynamics.

Inspired by biological vision mechanisms, event cameras~\citep{lichtsteiner2008128，zhang2023blink，gallego2020event，perot2020learning，mitrokhin2018event，gao2024sd2event} offer an alternative sensing paradigm by asynchronously capturing pixel-level brightness changes with microsecond-level temporal resolution and ultra-high dynamic range. These characteristics provide rich complementary information to RGB data, especially in high-frequency motion regions where conventional images are prone to blurring or overexposure. Consequently, a growing body of research has explored RGB-Event fusion frameworks, with methods such as AFNet~\citep{zhang2023frame}, VisEvent~\citep{wang2023visevent}, and CEUTrack~\citep{tang2022revisiting} integrating the two modalities via cross-modal attention, feature alignment, or spatiotemporal Transformers. However, most existing approaches operate predominantly in the spatial domain, overlooking the inherent frequency-domain relationship between RGB and event streams—a limitation that restricts the model's ability to fully exploit structural and motion information embedded in different frequency bands.

While Transformer-based fusion architectures have yielded promising results, they exhibit notable limitations: the self-attention mechanism~\citep{vaswani2017attention，dosovitskiy2020image} suffers from high computational complexity for long token sequences, making it inefficient for processing high-frequency event streams. More importantly, existing RGBE tracking methods generally lack dedicated frequency-domain modeling. From an imaging mechanism perspective, event cameras asynchronously respond to brightness differentials and their outputs densely encode rapid changes in the scene, thereby dominating high-frequency motion information. In contrast, RGB cameras capture absolute intensity through integration, emphasizing low-frequency structural content. This frequency-domain complementarity has yet to be effectively exploited. Furthermore, the static filter weight strategies in traditional frequency-domain methods are ill-suited to adapt to the sparse and dynamic characteristics of event data. Thus, accurately capturing fine motion boundaries and recovering clear textures under degraded conditions such as motion blur and low-light constitute a fundamental challenge for current methods.

To address these issues, we propose FreqTrack, a novel RGBE tracking framework that systematically incorporates frequency-domain transformations to achieve robust multimodal representation. Our approach is built upon two core innovations. First, we design a Spectral Enhancement Transformer (SET) layer, which introduces multi-head dynamic Fourier~\citep{cooley1965algorithm} filtering on top of standard multi-head self-attention. Each SET layer comprises 12 parallel one-dimensional frequency-domain filter heads, enabling adaptive optimization of token sequences through complex-valued weight modulation and routing mechanisms. This allows the model to selectively enhance cross-modal frequency components while suppressing noise. Second, we develop a Wavelet Edge Refinement (WER) module, which employs a learnable discrete wavelet transform based on Haar wavelets~\citep{mallat2002theory} to explicitly decompose event features into approximation and detail coefficients. Combined with convolutional layers and normalization, this module extracts multi-scale edge structures and supplies refined event features to each main network layer, significantly improving tracking robustness in motion blur and low-light scenarios.

Extensive experiments on two RGB-event tracking benchmarks, COESOT~\citep{tang2022revisiting} and FE108~\citep{zhang2021object}, demonstrate that FreqTrack achieves highly competitive performance. On the large-scale COESOT dataset, it attains state-of-the-art precision and strong success rate, while on the smaller FE108 dataset, it delivers competitive precision and reasonable success performance. These results validate the importance of frequency-domain modeling in RGBE tracking and highlight the effectiveness of the proposed modules.

In summary, our main contributions include:

\begin{enumerate}
\item We propose FreqTrack, a framework that incorporates frequency-domain modeling via dynamic spectral filtering and wavelet-based edge enhancement to effectively capture cross-modal complementary information.
\item We design the SET layer and WER module, which adaptively enhance cross-modal frequency features and extract multi-scale edge structures, respectively.
\item We validate FreqTrack on two benchmarks, COESOT and FE108. On COESOT, it achieves state-of-the-art precision and highly competitive overall performance, demonstrating strong generalization ability with sufficient training data.
\end{enumerate}

\section{Related Work}

\subsection{RGB-Event Tracking}

Research on RGB-Event (RGBE) fusion for object tracking has gained increasing attention, aiming to combine the rich texture of RGB images with the high dynamic response of event streams to address visual challenges in complex scenarios. ~\citet{zhang2021object} introduces a multi-modal alignment and fusion module to effectively integrate RGB and event data with different sampling rates, achieving robust tracking at high frame rates. The VisEvent~\citep{wang2023visevent} method constructs a comprehensive dataset containing 820 visible-event video pairs and establishes a baseline using a cross-modality Transformer (CMT) to enhance feature interaction. CEUTrack~\citep{tang2022revisiting} unifies RGB frames and color-event voxels through a single-stage backbone, enabling simultaneous feature extraction, matching, and interactive learning. AFNet~\citep{zhang2023frame} further incorporates an event-guided cross-modal alignment (ECA) module and cross-correlation fusion (CF) to improve target localization in dynamic environments.   ~\citet{zhu2023cross} proposed a masked modeling strategy that randomly masks tokens of modalities to bridge the distribution gap between RGB and event data, thereby enhancing model generalization. HDETrack~\citep{wang2024event} pioneers the application of knowledge distillation in multimodal tracking, transferring multi-view (event image-voxel) knowledge to single-modality event tracking and expanding the utility boundaries of event data.

\subsection{Frequency domain learning}
\begin{figure*}
  \centering
  \includegraphics[width=1.0\linewidth]{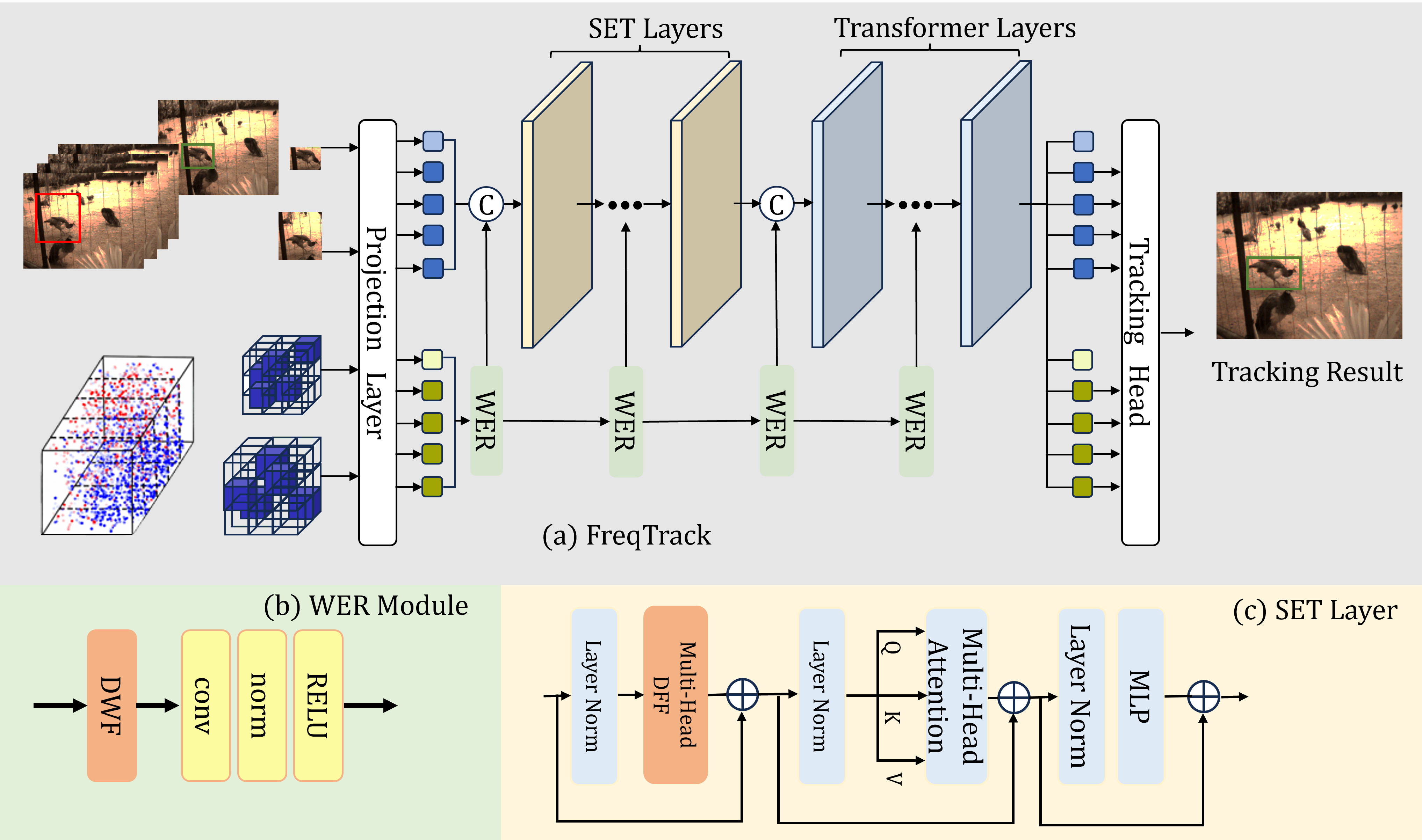}
   \caption{The FreqTrack network framework and core modules. (a) Overall pipeline from input to output. (b) Wavelet Edge Refinement (WER) module employing Dynamic Wavelet Filtering (DWF). (c) Spectral Enhancement Transformer (SET) layer with Dynamic Fourier Filtering (DFF).}
   \label{fig1:FreqTrack}
\end{figure*}
Frequency-domain analysis has long served as a fundamental tool in signal and image processing. Early work such as the Fast Fourier Transform (FFT) laid the groundwork for efficient spectral analysis~\citep{cooley2007fast}, while subsequently developed adaptive filtering techniques achieved noise suppression while preserving structural details such as edges~\citep{deng1993adaptive}. The capacity of wavelet transforms to decompose signals locally into multi-scale representations has led to their widespread adoption in enhancement tasks such as medical imaging ~\citep{yang2010medical}.

With the rise of deep learning, frequency-domain reasoning has regained prominence in modern architectures. Recent studies have revealed intriguing properties of Vision Transformers (ViTs) in the frequency domain: research in ~\citet{park2022vision} and ~\citet{si2022inception} demonstrates that the self-attention mechanism in ViTs tends to capture low-frequency global structures but struggles with high-frequency local details. This finding is critical, as high-frequency information is often indispensable for precise pixel-level prediction tasks. Studies in ~\citet{chen2024semantic} show that downsampling operations can cause high-frequency aliasing, adversely affecting edge quality in semantic segmentation, whereas ~\citet{cai2021frequency} successfully leverages frequency constraints to achieve more realistic and identity-preserving image translation.

Frequency-domain guidance has also proven effective in image restoration and enhancement tasks. In image dehazing, ~\citet{yu2022frequency} combines frequency and spatial cues to recover clearer structures. Similarly, research indicates that diffusion models exhibit biases in frequency components, which can be exploited to improve generation outcomes ~\citep{everaert2024exploiting}. Even in classical problems such as image completion, frequency characteristics have been implicitly utilized to maintain structural consistency ~\citep{drori2003fragment}.

Notably, the integration of frequency-domain learning into RGB-Event tracking has seen preliminary exploration. FAEFTrack~\cite{shang2025improving} introduced a multi-frequency attention mechanism that maps different frequency components into channel dimensions via Discrete Cosine Transform and processes them through channel attention. However, this approach essentially constitutes a channel-wise frequency recombination strategy; while it distributes frequency information across channels, it falls short of achieving active selection and filtering of specific frequency components within the frequency domain. To address this limitation, we propose the FreqTrack framework, whose core innovation lies in introducing dynamic frequency-domain filters. This design enables the network to directly suppress interference from irrelevant frequency bands during forward propagation while proactively enhancing the extraction of critical frequency features, thereby achieving more precise and efficient feature enhancement in the frequency domain.

\section{Our Proposed Approach}
\subsection{Overview}
This section presents FreqTrack, a unified frequency-learning framework for RGB-Event object tracking. The core idea incorporates dynamic frequency-domain modeling into the Transformer architecture to fully exploit cross-modal complementary information. Built upon a ViT backbone, our approach integrates two key components in each layer: the Spectral Enhancement Transformer (SET) layer, which augments self-attention with multi-head dynamic Fourier filtering to adaptively modulate frequency components, and the Wavelet Edge Refinement (WER) module, which leverages learnable wavelet transforms to extract multi-scale edge structures from event data and progressively refines event representations through layer-wise interaction with the main network. The overall architecture is shown in Fig~\ref{fig1:FreqTrack}, and the following sections detail each component.

\begin{figure}[t]
  \centering
  \includegraphics[width=0.8\linewidth]{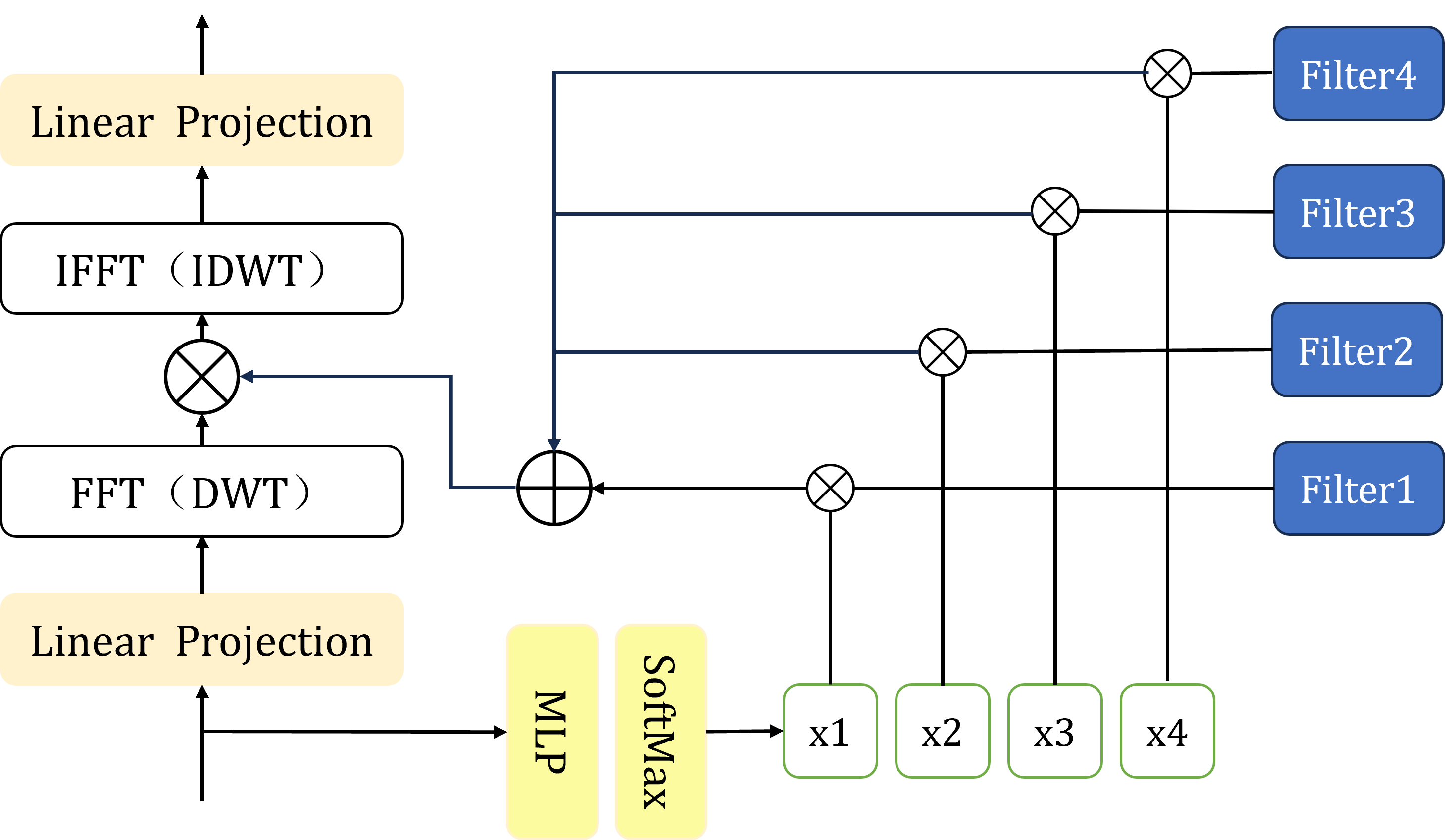}
   \caption{Structure of the Dynamic Filters (DFF/DWF). Dynamic Fourier Filtering(DFF) and Dynamic Wavelet Filtering(DWF) share the same adaptive filtering architecture but differ in their underlying transformations: Fourier and Wavelet transforms, respectively}
   \label{fig2:dff}
\end{figure}

\subsection{Input Representation}
Our framework adopts RGB video frames and asynchronous event streams as dual-modal input, achieving spatiotemporal consistency across modalities through dynamic temporal alignment and structured encoding.

For RGB frame,the RGB video sequence is denoted as $\mathcal{I} = \{I_1, I_2, \ldots, I_N\}$  where $I_i$ represents the i-th RGB frame, and N is the total number of frames. We crop the template patch $Z^I$ (target initialization region) and search patch $X^I$ (candidate search region) from RGB frames.

For event stream, the asynchronous event stream is represented as ${\mathcal{E}}=\{e_{j}=(x_{j},y_{j},t_{j},p_{j})\}_{j=1}^{M}$,where each event $e_j$ contains pixel coordinates $(x_j, y_j)$,timestamp $t_j$ and polarity $ p_j \in \{ +1, -1 \} $,with M being the total number of events. To achieve temporal alignment with RGB frames, we adopt the time surface representation~\citep{lagorce2016hots} of events. For each RGB frame timestamp $t_{RGB}^{(i)}$,we construct an event window covering its exposure duration and compute event contributions via linear interpolation:
\begin{equation}
  V_t(x, y) = \sum_j p_j \cdot \max\left(0, 1 - \frac{|t_{\mathrm{RGB}}^{(i)} - t_j|}{\Delta t}\right)
  \label{eq:volumetric_representation}
\end{equation}
where ${\Delta t}$ is the exposure time of RGB frames, the weight linearly decays with the temporal distance between events and RGB timestamps.

\subsection{Backbone Network}
Before feeding multimodal data into the Transformer network, we transform them into token sequence representations through a projection layer. This layer comprises four independent convolutional branches dedicated to different input sources. For the RGB modality, the template region ${\cal P}_t^{\mathrm{RGB}}$ and search region ${\cal P}_s^{\mathrm{RGB}}$ are projected into feature embeddings ${\cal E}_t^{\mathrm{RGB}} \in \mathbb{R}^{N_t \times D}$ and ${\cal E}_s^{\mathrm{RGB}} \in \mathbb{R}^{N_s \times D}$ via $16\times16$ convolutional operations. For the event modality, the template voxels ${\cal V}_t^{\mathrm{EV}}$ and search voxels ${\cal V}_s^{\mathrm{EV}}$ are projected into ${\cal E}_t^{\mathrm{EV}} \in \mathbb{R}^{N_t \times D}$ and ${\cal E}_s^{\mathrm{EV}} \in \mathbb{R}^{N_s \times D}$ through $4\times4$ convolutional operations. All tokens are augmented with learnable position embeddings, where template regions share position encoding ${\cal P}_t \in \mathbb{R}^{N_t \times D}$ and search regions share ${\cal P}_s \in \mathbb{R}^{N_s \times D}$.

After obtaining the projected tokens, the template and search tokens from the event modality are first concatenated:
\begin{equation}
{\cal E}^{\mathrm{EV}} = [{\cal E}_t^{\mathrm{EV}}; {\cal E}_s^{\mathrm{EV}}] \in \mathbb{R}^{(N_t+N_s) \times D}
\end{equation}

This concatenated event sequence is then fed as a whole into the Wavelet Edge Refinement module for feature enhancement:
\begin{equation}
\tilde{{\cal E}}^{\mathrm{EV}} = \mathrm{WER}({\cal E}^{\mathrm{EV}})
\end{equation}

The enhanced event tokens are reorganized with RGB tokens to form a unified sequence:
\begin{equation}
{\cal H}_0 = [{\cal E}_t^{\mathrm{RGB}}; \tilde{{\cal E}}_t^{\mathrm{EV}}; {\cal E}_s^{\mathrm{RGB}}; \tilde{{\cal E}}_s^{\mathrm{EV}}] \in \mathbb{R}^{2(N_t+N_s) \times D}
\end{equation}
where $\tilde{{\cal E}}_t^{\mathrm{EV}}$ and $\tilde{{\cal E}}_s^{\mathrm{EV}}$ are split from $\tilde{{\cal E}}^{\mathrm{EV}}$.

As shown in Fig.~\ref{fig1:FreqTrack}, our backbone network consists of 12 layers, with the first 6 layers employing Spectral Enhancement Transformer (SET) layers, as shown in Fig.~\ref{fig1:FreqTrack}(c), and the latter 6 layers using standard Transformer layers. The SET layer incorporates Dynamic Fourier Filtering based on the standard Transformer:
\begin{align}
{\cal H}''_l &= {\cal H}'_l + \mathrm{DFF}(\mathrm{LN}({\cal H}'_l)) \\
{\cal H}'_l &= {\cal H}_l + \mathrm{MSA}(\mathrm{LN}({\cal H}_l)) \\
{\cal H}_{l+1} &= {\cal H}''_l + \mathrm{MLP}(\mathrm{LN}({\cal H}''_l))
\end{align}

The subsequent 6 standard Transformer layers follow the classical design:
\begin{align}
{\cal H}'_l &= {\cal H}_l + \mathrm{MSA}(\mathrm{LN}({\cal H}_l)) \\
{\cal H}_{l+1} &= {\cal H}'_l + \mathrm{MLP}(\mathrm{LN}({\cal H}'_l))
\end{align}
Unlike two-stream Siamese trackers that use Transformer layers solely for sensor data fusion, our approach enables adaptive data fusion while preserving discriminative information by directly concatenating template and search region tokens from both sensor modalities.

\subsection{Spectral Enhancement Transformer Layer}
\textbf{Dynamic Fourier Filtering Mechanism.} The Spectral Enhancement Transformer (SET) layer is designed to address core challenges in RGBE tracking, where target appearance often undergoes complex variations due to occlusion, illumination changes, and dynamic backgrounds. While event data provides high-temporal-resolution edge structures, it remains sparse and noise-sensitive. Traditional spatial attention mechanisms face limitations in handling such issues, often struggling to capture stable structural patterns shared across modalities, particularly those more evident in the frequency domain. To this end, building on prior work in frequency-domain adaptive processing~\citep{tatsunami2024fft}, we introduce the Dynamic Fourier Filtering (DFF) mechanism as a structural saliency encoder, enabling the network to develop frequency preference capabilities crucial for robust tracking.The structure of the Dynamic Fourier Filtering (DFF) mechanism is shown in Fig.~\ref{fig2:dff}.

The mathematical formulation begins with an input token sequence $\mathbf{H} \in \mathbb{R}^{N \times D}$, which first undergoes linear projection to an intermediate representation:
\begin{equation}
\mathbf{H}' = \mathbf{W}_1 \mathbf{H} + \mathbf{b}_1
\end{equation}
Application of 1D Discrete Fourier Transform along the sequence dimension yields frequency-domain representations where low-frequency components correspond to stable shape-related structural patterns, while high-frequency components reflect edges, rapid motion, and event stream activations. The key innovation lies in the dynamic filtering strategy—different tracking scenarios require distinct frequency preferences: fast-motion scenarios necessitate enhanced high-frequency responses from events, sudden illumination changes demand suppression of high-frequency noise, and stable scenes benefit from boosted low-frequency structural patterns from RGB. This adaptability is achieved through input-dependent routing weights:
\begin{equation}
\alpha = \mathrm{Softmax}\left(\mathrm{MLP}\left(\frac{1}{N}\sum_{i=1}^{N} \mathbf{H}_i\right)\right) \in \mathbb{R}^{K}
\end{equation}
These weights combine $K$ learnable basis filters for frequency-selective modulation:
\begin{equation}
\tilde{\mathbf{H}}_{\text{freq}} = \mathbf{H}_{\text{freq}} \odot \sum_{k=1}^{K} \alpha_k \cdot \mathbf{W}_k
\end{equation}
The final output is recovered through inverse transformation:
\begin{equation}
\tilde{\mathbf{H}} = \mathbf{W}2 \cdot \mathcal{F}^{-1}(\tilde{\mathbf{H}}_{\text{freq}}) + \mathbf{b}_2
\end{equation}
This design provides significant advantages for RGBE tracking by filtering out background variations irrelevant to tracking in the frequency domain prior to spatial attention, enhancing high-frequency components representing motion boundaries in event streams, strengthening low-frequency structures related to target shape in RGB data, and generating scene-adaptive filtering preferences driven by input characteristics.

\textbf{Spectral Enhancement Transformer Layer.} The SET layer, as shown in Fig.~\ref{fig1:FreqTrack}(c), adopts a "frequency-first, spatial-later" processing paradigm that coordinates spectral and spatial optimization. The architecture processes input tokens through three sequential operations while maintaining residual connections throughout. The computational flow begins with spectral enhancement via dynamic Fourier filtering:

\begin{equation}
\mathbf{H}'_l = \mathbf{H}_l + \mathrm{DFF}(\mathrm{LN}(\mathbf{H}_l))
\end{equation}

This initial frequency-domain operation, implemented through a multi-head dynamic filtering architecture that parallels the attention mechanism, serves as structural prior enhancement, helping to filter noise and improve structural consistency before spatial modeling. The frequency-enhanced features subsequently undergo spatial context modeling through multi-head self-attention:

\begin{equation}
\tilde{\mathbf{H}}_l = \mathbf{H}'_l + \mathrm{MSA}(\mathrm{LN}(\mathbf{H}'_l))
\end{equation}

At this stage, the attention mechanism more readily locks onto target regions because the input has been purified in the frequency domain—irrelevant background variations are suppressed while task-relevant structural patterns are enhanced. Finally, channel-wise transformation and feature refinement complete the processing:

\begin{equation}
\mathbf{H}_{l+1} = \tilde{\mathbf{H}}_l + \mathrm{MLP}(\mathrm{LN}(\tilde{\mathbf{H}}_l))
\end{equation}

The MLP reorganizes channel features to properly integrate frequency-domain enhancements into network representations. We deploy SET layers in the front part of the network to enhance shallow feature representations, while standard Transformer layers are employed in the later stages to focus on high-level semantic information. This layered design provides valuable frequency-dimensional reasoning capabilities for multimodal tracking tasks while maintaining computational efficiency without increasing model complexity.

\subsection{Wavelet Edge Refinement Module}The Wavelet Edge Refinement (WER) module is designed to address the unique characteristics of event data through wavelet transform, which provides superior localization capabilities for analyzing transient signals compared to Fourier transforms. The detailed structure of the Wavelet Edge Refinement (WER) module is illustrated in Fig.~\ref{fig1:FreqTrack}(b).The module employs Dynamic Wavelet Filtering (DWF), as shown in Fig.~\ref{fig2:dff}, to adaptively process event features in the wavelet domain. While event streams inherently encode edge information and motion boundaries through pixel-level brightness changes, these features exhibit strong localization in both spatial and temporal domains. The Discrete Wavelet Transform (DWT) offers an effective mathematical framework for such analysis by decomposing signals into approximation coefficients that capture low-frequency trends and detail coefficients that preserve high-frequency information. For an input signal $x[n]$, the transform is computed as:
\begin{equation}
cA[k] = \sum_n x[n] \cdot \phi_{j,k}[n], \quad cD[k] = \sum_n x[n] \cdot \psi_{j,k}[n]
\end{equation}
where $cA[k]$ and $cD[k]$ represent approximation and detail coefficients respectively, with $\phi_{j,k}$ and $\psi_{j,k}$ being the scaling and wavelet functions.

In our architecture, the WER module operates synchronously with each main network layer, establishing a dedicated processing pathway for event data refinement. The module receives concatenated event token sequences from both template and search regions, formulated as $\mathbf{E}^{ev} = [\mathbf{E}_t^{ev}; \mathbf{E}_s^{ev}] \in \mathbb{R}^{(N_t+N_s) \times D}$. The processing pipeline employs learnable Haar wavelets to decompose the input event tokens through DWT, followed by dynamic filtering operations in the wavelet domain. The forward propagation begins with:
\begin{equation}
\mathbf{cA}, \mathbf{cD} = \mathrm{DWT}(\mathbf{E}^{ev})
\end{equation}
where $\mathbf{cA}$ captures approximate low-frequency components of event activity while $\mathbf{cD}$ isolates detailed high-frequency edge information. Dynamic filtering is then applied separately to these coefficients:
\begin{equation}
\tilde{\mathbf{cA}} = \mathbf{cA} \odot \mathbf{W}{cA}, \quad \tilde{\mathbf{cD}} = \mathbf{cD} \odot \mathbf{W}{cD}
\end{equation}
The filtered coefficients undergo reconstruction via inverse wavelet transform:
\begin{equation}
\tilde{\mathbf{E}}^{ev} = \mathrm{IDWT}(\tilde{\mathbf{cA}}, \tilde{\mathbf{cD}})
\end{equation}
This refined event representation is subsequently split back into template and search components $\tilde{\mathbf{E}}_t^{ev}$ and $\tilde{\mathbf{E}}_s^{ev}$ for cross-modal integration in the main network. The modular design ensures progressive refinement of event features across network depths, with each WER instance processing its predecessor's output to build increasingly sophisticated representations while maintaining temporal precision.

\subsection{Head and Loss Function}

We adopt a widely adopted tracking head, which generates three distinct outputs: a classification score map for target-background discrimination, a bounding box size map, and a local offset map. For model optimization, we utilize a composite loss function combining focal loss~\citep{lin2017focal} for classification, L1 loss~\citep{girshick2015fast} for bounding box regression, and GIoU loss~\citep{rezatofighi2019generalized} for spatial alignment. The overall objective function is formulated as:
\begin{equation}
L = \lambda_{focal}L_{focal} + \lambda_{1}L_{1} + \lambda_{GIoU}L_{GIoU}
\end{equation}
where $\lambda_{focal} = 1$, $\lambda_{1} = 14$ and $\lambda_{GIoU} = 1$ are the hyperparameters in our experiment.

\section{Experiment}
\subsection{Experimental Settings}
\textbf{Implementation Details.} We implemented the proposed FreqTrack framework using PyTorch and trained it on 2 NVIDIA RTX 4090 GPUs. Specifically, we adopted the AdamW~\citep{loshchilov2017decoupled} optimizer, with the learning rate, batch size, and weight decay set to 0.0001, 8, and 0.0001, respectively. The learning rate scheduling followed a StepLR strategy with a decay rate of 0.1. For the backbone, we employed a lightweight pre-trained Vision Transformer model.

\textbf{Evaluation metrics.} In our experimental evaluation, tracking performance is examined through three widely used metrics: Success Rate (SR) and Precision Rate (PR). SR represents the fraction of frames where the Intersection over Union (IoU) between the predicted and ground-truth bounding boxes exceeds a given threshold, indicating the robustness of the tracker. PR captures the proportion of frames in which the distance between the predicted and ground-truth target centers falls below a specified pixel radius, serving as an indicator of localization accuracy.

\textbf{Dataset.} We evaluate the proposed method on two large-scale RGB-Event tracking datasets: FE108~\citep{zhang2021object} and COESOT~\citep{tang2022revisiting}. The FE108 dataset is a multimodal visual tracking benchmark composed of synchronized RGB frames and event streams across 108 sequences covering diverse scenes and object categories, designed to facilitate the evaluation of tracker robustness under challenging conditions such as rapid motion, illumination variation, and dynamic backgrounds.The COESOT dataset is a large-scale RGB-Event collaborative tracking benchmark featuring over ninety object categories and more than a thousand paired sequences, providing a comprehensive and systematic platform for assessing high-performance tracking algorithms that integrate color and event-based information.

\subsection{Comparisons with State-of-the-arts}

Our experiments are conducted on two frame-event tracking benchmarks: FE108 and COESOT dataset.
\begin{table}[t]
\caption{Tracking results on COESOT dataset}
\label{tab1:coesot}
\centering
\begin{tabular}{l|l|l}
\hline
\textbf{Method} & \textbf{SR(\%)} & \textbf{PR(\%)}  \\
\hline
EventTPT~\citep{xia2025towards} & \textbf{64.7} & 73.0  \\
FAFETrack~\citep{shang2025improving} & 62.5 & 76.5 \\
CEUTrack~\citep{tang2022revisiting} & 62.7 & 76.0\\
TransT~\citep{chen2021transformer} & 60.5 & 72.4 \\
TrDiMP~\citep{wang2021transformer} & 60.1 & 72.2 \\
KeepTrack~\citep{mayer2021learning} & 59.6 & 70.9 \\
OSTrack~\citep{ye2022joint} & 59.0 & 70.7 \\
AiATrack~\citep{gao2022aiatrack} & 59.0 & 72.4 \\
DiMP50~\citep{bhat2019learning} & 58.9 & 72.0 \\
STARK-S50~\citep{yan2021learning} & 55.7 & 66.7 \\
MixFormer22k~\citep{cui2023mixformerv2} & 55.7 & 66.3 \\
ATOM~\citep{danelljan2019atom} & 55.0 & 68.8 \\
SiamRPN~\citep{li2018high} & 53.5 & 65.7 \\
\hline
\textbf{FreqTrack} & 62.7 & \textbf{76.6} \\
\hline
\end{tabular}
\end{table}
\begin{figure}[t]
  \centering
  \includegraphics[width=0.8\linewidth]{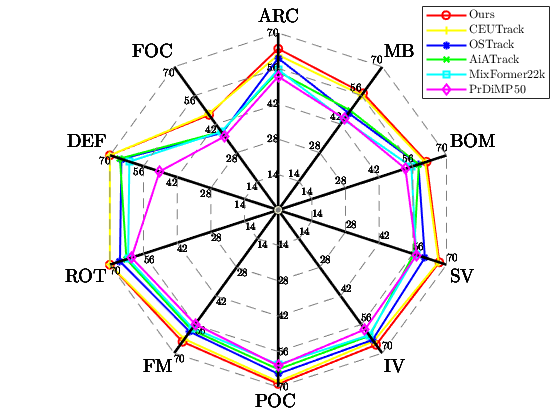}
   \caption{Comparison of success rates under different challenging scenarios on COESOT.}
   \label{fig3:radar}
\end{figure}

\textbf{Results on COESOT dataset.} Table \ref{tab1:coesot} presents the tracking results of the proposed method on the COESOT dataset. Our approach achieves a Success Rate (SR) of \textbf{62.7\%} and a Precision Rate (PR) of \textbf{76.6\%}, demonstrating highly competitive overall performance on this benchmark.Specifically, our method exhibits outstanding performance in terms of Precision Rate (PR), achieving the highest value among all compared methods. Compared to the recently proposed FAFETrack (SR 62.5\%, PR 76.5\%), we achieve a slight improvement in both success and precision rates. Furthermore, when compared to EventTPT (SR 64.7\%, PR 73.0\%), our method achieves a notable \textbf{+3.6\%} improvement in precision while maintaining a competitive success rate. These consistent results validate the effectiveness of our proposed approach.

For a fine-grained analysis under specific challenging conditions, we present the success rate radar chart in Fig.~\ref{fig3:radar}. The results across ten challenging scenarios show that our FreqTrack achieves leading or highly competitive performance in most cases. Notably, it attains the highest success rates in scenarios such as deformation (DEF), rotation (ROT), fast motion (FM), illumination variation (IV), and motion blur (MB). This consistent advantage across various challenging scenarios, particularly in addressing key difficulties like dynamic appearance changes and motion blur, demonstrates the robustness of our frequency-domain modeling approach in enhancing feature representation and motion reasoning. These consistent results validate the effectiveness of our proposed approach.

\begin{table}[t]
\caption{Tracking results on FE108 dataset}
\label{tab2:fe108}
\centering
\begin{tabular}{l|l|l}
\hline
\textbf{Method} & \textbf{SR(\%)} & \textbf{PR(\%)}  \\
\hline
CMT-ATOM~\citep{wang2023visevent} & \textbf{54.3} & 79.4  \\
PrDiMP~\citep{danelljan2020probabilistic} & 53.0 & \textbf{80.5} \\
DiMP~\citep{bhat2019learning} & 52.6 & 79.1 \\
ATOM~\citep{danelljan2019atom} & 46.5 & 71.3 \\
SiamRPN~\citep{li2018high} & 21.8 & 33.5 \\
SiamFC++~\citep{xu2020siamfc++} & 23.8 & 39.1 \\
KYS~\citep{bhat2020know} & 26.6 & 41.0 \\
\hline
\textbf{FreqTrack} & 49.7 & 79.1 \\
\hline
\end{tabular}
\end{table}
\textbf{Results on FE108 dataset.} Table \ref{tab2:fe108} presents the tracking results of the proposed FreqTrack on the FE108 dataset. Our method achieves a SR of 49.7\% and a Precision Rate (PR) of 79.1\%. On the relatively small-scale FE108 dataset, our method demonstrates competitive performance in PR, which is on par with the top-performing method. However, a performance gap in SR is observed compared to other state-of-the-art methods. We attribute this primarily to the limited scale of FE108, which contains only 76 training sequences, potentially restricting the model's ability to fully learn robustness across diverse and challenging scenarios. In contrast, on the large-scale COESOT dataset comprising 827 training sequences, our method achieves more competitive results (SR 62.7\%, PR 76.6\%), validating its capacity to learn effective representations from sufficient data.

\subsection{Ablation Study}
\begin{table}[t]
\caption{Ablation study for important components on COESOT dataset. $\times$ represents the component is removed.}
\label{tab3:ablation_study}
\centering
\begin{tabular}{c|cccc|cc}
\hline
\textbf{\#} & \textbf{RGB} & \textbf{Event} & \textbf{SET}& \textbf{WER}& \textbf{SR(\%)}& \textbf{PR(\%)}\\
\hline
1& $\times$ & & & & 40.1 & 48.2 \\
2& & $\times$ & & & 57.1 & 72.3 \\
3& & & $\times$ & $\times$ & 61.6 & 75.2 \\
4& & & & $\times$ & 62.4 & 76.1 \\
\hline
5& & & & & \textbf{62.7} & \textbf{76.6} \\
\hline
\end{tabular}
\end{table}
\textbf{Impact of Multimodal Input.} To comprehensively evaluate the contribution of multimodal input to tracking performance, we conduct ablation experiments under different modality configurations. As demonstrated in Table \ref{tab3:ablation_study} (rows 1, 2, and 5), the model using only Event data achieves a success rate (SR) of 57.1\% and a precision rate (PR) of 72.3\%. In contrast, the model relying solely on RGB input yields significantly lower performance, with only 40.1\% SR and 48.2\% PR. This substantial performance gap underscores the critical importance of event data in maintaining tracking robustness. The full model, which effectively integrates both RGB and Event modalities, achieves the best performance (62.7\% SR, 76.6\% PR), demonstrating that our fusion framework successfully leverages the complementary strengths of both modalities: RGB provides rich spatial appearance information while Event data delivers superior temporal resolution for motion tracking.

\textbf{Effectiveness of the Spectral Enhancement Transformer Layer.} To validate the contribution of the Spectral Enhancement Transformer (SET) layer, we analyze its impact in conjunction with the Wavelet Edge Refinement module. As shown in Table \ref{tab3:ablation_study} (rows 3 and 5), when both the SET layer and WER module are removed, the performance drops to 61.6\% SR and 75.2\% PR. Compared to the complete model (62.7\% SR, 76.6\% PR), this represents a noticeable degradation in both metrics. The performance decline confirms SET's effectiveness in enhancing feature representations through frequency-domain processing. By adaptively modulating frequency components via dynamic Fourier filtering, the SET layer enables the model to better capture structural patterns and suppress noise interference, thereby improving tracking accuracy in challenging scenarios with appearance variations and motion blur.

\textbf{Effectiveness of the Wavelet Edge Refinement Module.} We further investigate the specific contribution of the Wavelet Edge Refinement (WER) module to the overall tracking performance. As reported in Table \ref{tab3:ablation_study} (rows 4 and 5), removing only the WER module causes performance to decrease from 62.7\% to 62.4\% in SR and from 76.6\% to 76.1\% in PR. Though the performance drop is modest, the consistent degradation across both metrics indicates WER's positive role in refining event representations. The module employs learnable wavelet transforms and dynamic wavelet filtering (DWF) to extract multi-scale edge features from event data, effectively enhancing structural information while suppressing noise. The results confirm that the progressive refinement of event features through WER's layer-wise processing contributes to more discriminative representations for accurate target localization.
\begin{figure}[t]
  \centering
  \includegraphics[width=1.0\linewidth]{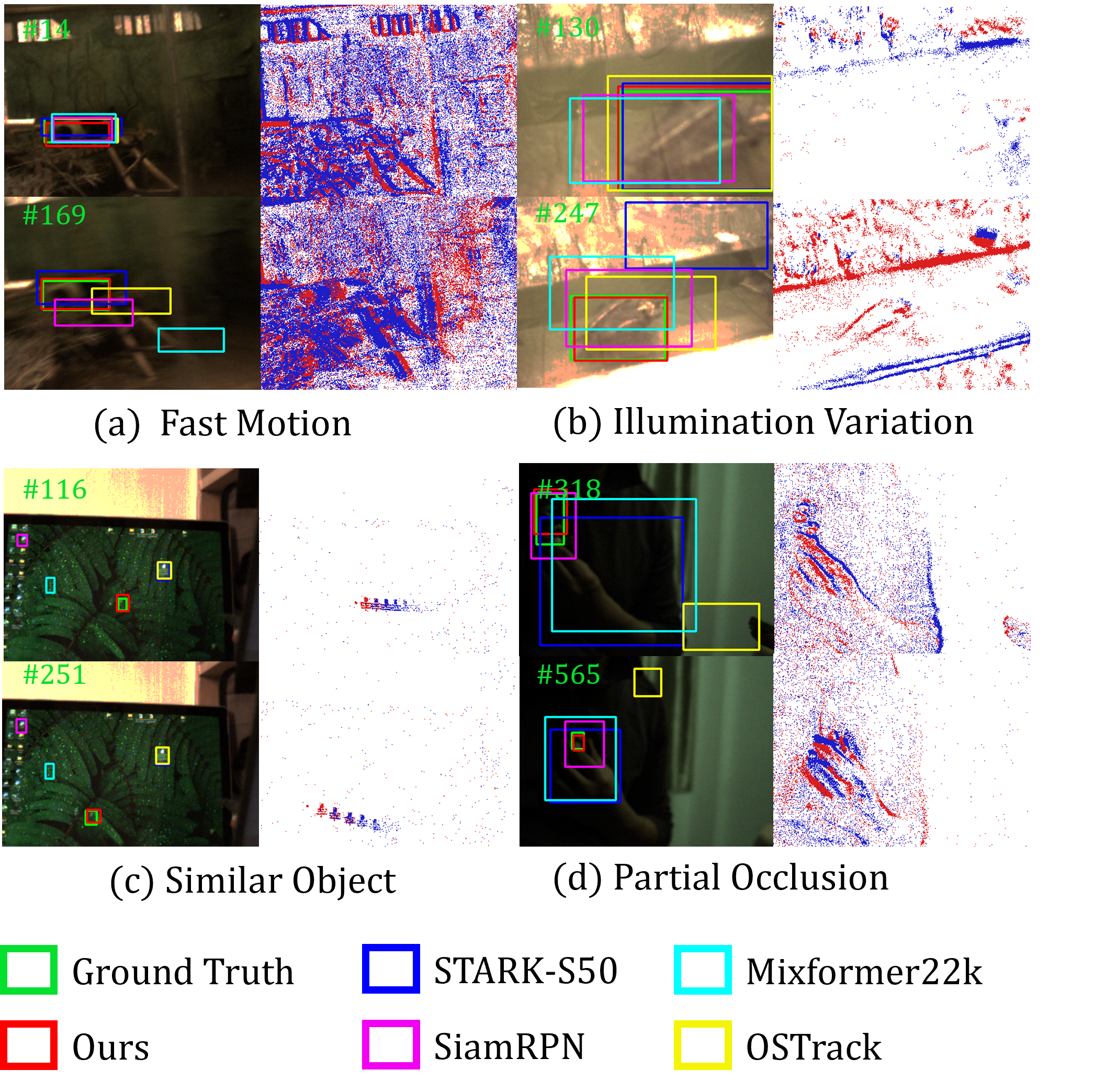}
   \caption{Tracking results of our FreqTrack under 4 challenging scenarios on COESOT.Event images are used for visual comparison only.}
   \label{fig4:visual}
\end{figure}

\subsection{Visualization}
To comprehensively evaluate the tracking performance of FreqTrack in complex scenarios, we selected four representative cases from the 17 challenging scenarios in the COESOT dataset for visualization analysis, including fast motion, illumination variation, similar object interference, and partial occlusion. As shown in Figure 4, we compare the ground truth bounding boxes with the predictions from our method and four baseline methods.

The visualization results demonstrate that FreqTrack achieves excellent performance across the selected challenging conditions, reflecting its overall competitiveness in a broader range of challenging scenarios. Specifically, in fast motion scenarios, our method effectively captures motion boundaries through frequency-domain modeling, overcoming positioning drift caused by motion blur. Under severe illumination changes, the robustness of frequency-domain features to lighting variations enables continuous target tracking. When facing similar object interference, the discriminative capability of spectral features helps distinguish between the target and distractors. In partial occlusion cases, the multi-scale edge information extracted by the Wavelet Edge Refinement module helps maintain reasonable estimation of the target structure.

These representative cases validate the general advantages of frequency-domain modeling: the Spectral Enhancement Transformer (SET) layer enhances the extraction of motion information and global structures, while the Wavelet Edge Refinement (WER) module preserves critical local details. Although only four representative scenarios are shown here, selected from the complete range of challenges in the dataset, they sufficiently demonstrate the robustness and generalization capability of FreqTrack in complex real-world scenarios.

\section{Conclusion}
Based on the comprehensive analysis of existing RGB-Event tracking methods, this paper presents FreqTrack, an effective tracking framework that incorporates frequency-domain modeling through dynamic spectral filtering and wavelet-based edge enhancement. Specifically, the Spectral Enhancement Transformer (SET) layer enables adaptive frequency component selection via dynamic Fourier filtering, while the Wavelet Edge Refinement (WER) module extracts multi-scale edge structures from event streams. Extensive experiments conducted on COESOT and FE108 benchmarks demonstrate the competitive performance of FreqTrack in both precision and robustness. Ablation studies and qualitative visualizations further validate the effectiveness of each proposed component in enhancing feature representation and motion reasoning. Benefiting from its frequency-aware architecture, FreqTrack exhibits strong potential for handling challenging scenarios involving motion blur and illumination variations. In future work, we plan to explore more efficient frequency-domain fusion strategies and extend the framework to other multimodal vision tasks.

\bibliography{iclr2026_conference}
\bibliographystyle{iclr2026_conference}

\end{document}

%% file: math_commands.tex
\usepackage{amsmath,amsfonts,bm}

\def\eqref#1{equation~\ref{#1}}

\def\1{\bm{1}}

\DeclareMathAlphabet{\mathsfit}{\encodingdefault}{\sfdefault}{m}{sl}
\SetMathAlphabet{\mathsfit}{bold}{\encodingdefault}{\sfdefault}{bx}{n}

